\documentclass{article}

\usepackage[preprint]{corl_2026} 

\title{V2P-Manip: Learning Dexterous Manipulation from Monocular Human Videos}

\usepackage{graphicx}
\usepackage{amssymb}
\usepackage{amsmath}
\usepackage{makecell}
\usepackage{pifont}
\usepackage{bbm}
\usepackage{xcolor}
\usepackage{tabularx}
\usepackage{booktabs}
\author{%
  Kaihan Chen$^{1}$ \quad 
  Yanming Shao$^{2,3}$ \quad 
  Haifeng Ji$^{1}$ \quad 
  Xiaokang Yang$^{3}$ \quad 
  Yao Mu$^{2,3}$ \\[0.5ex]
  \textnormal{$^{1}$Zhejiang University, Hangzhou, China} \\
  \textnormal{$^{2}$Shanghai Jiao Tong University, Shanghai, China} \\
  \textnormal{$^{3}$Shanghai AI Laboratory, Shanghai, China} \\
}

\begin{document}
\maketitle


\begin{abstract}
Achieving autonomous robotic dexterous manipulation requires precise, human-like action sequences at scale. As a scalable supplement to costly teleoperation data, extracting trajectories with both visual fidelity and physical plausibility from monocular videos represents a promising frontier in embodied AI. To this end, we introduce V2P-Manip, an efficient framework designed to learn dexterous manipulation policies directly from human demonstration videos. We establish an efficient, integrated pipeline encompassing 3D asset acquisition, trajectory estimation, and dexterous policy learning. To bridge the gap between visual perception and physical constraints, we introduce a two-stage refinement process to enforce spatial alignment and physical consistency. Evaluations on the TACO and OakInk benchmarks demonstrate that our approach significantly outperforms previous methods in pose accuracy, adaptability to unstructured environments, and training efficiency. Ultimately, experimental results confirm an average success rate of over 75\% across multiple synthetic manipulation tasks and validate the adaptability of the extracted manipulation priors across diverse dexterous hand embodiments.
\end{abstract}

 
\section{Introduction}
\label{sec:intro}
\noindent Endowing robots with human-like dexterous manipulation capabilities is a long-standing goal in robotics. To achieve this, learning from human demonstrations \citep{argall2009survey,mccarthy2025towards,shao2021concept2robot,mandlekar2023mimicgen,jiang2025dexmimicgen} has emerged as a predominant paradigm. Traditionally, high-fidelity trajectory data are acquired through teleoperation \citep{ding2025bunny,zhu2020bimanual,gao2023dual} or optical motion capture systems \citep{liu2025forcemimic,fu2024mobile,zordan2003mapping}. However, these methods are often constrained by expensive hardware requirements, time-consuming calibration processes, and the limited diversity of captured scenarios, which hinders the scalability of robotic skill acquisition. In contrast, the vast amount of human manipulation videos available on the internet provides a rich and diverse source of demonstrations. Leveraging these video resources to learn manipulation skills offers a more scalable and promising alternative.

\medskip
\noindent While leveraging human demonstrations is highly promising, extracting executable trajectories from monocular videos presents formidable challenges. Independent hand and object estimators \citep{doosti2020hope,hamer2009tracking,armagan2020measuring} frequently suffer from spatial inconsistencies, primarily due to the inherent depth ambiguity and occlusion in 2D observations. These perceptual inaccuracies lead to physically implausible interactions such as interpenetration or unrealistic hovering \citep{grady2021contactopt}, which are detrimental to downstream policy training. Furthermore, acquiring robust manipulation skills requires navigating high-dimensional action spaces \citep{liu2023safe,lippi2020latent}, coupled with the need to address substantial noise in human priors. Traditional imitation learning often struggles with the morphology and domain gaps between humans and robots, whereas reinforcement learning \cite{kaelbling1996reinforcement} from scratch faces prohibitive exploration complexity in contact-rich manipulation tasks.

\medskip
\noindent To address these challenges, we present an efficient framework for acquiring dexterous skills directly from unstructured monocular RGB videos. By eliminating the necessity for specialized hardware,  our approach significantly lowers the barriers for large-scale data acquisition. We establish a comprehensive pipeline that transforms raw monocular observations into high-quality robotic demonstrations through a hierarchical two-stage optimization process. This refinement systematically rectifies spatial misalignments and enforces contact-level constraints, thereby ensuring both geometric alignment and physical feasibility in the resulting trajectories. Building upon these refined motion priors, we employ a residual learning scheme \cite{he2016deep,fang2021deep,shafiq2022deep} that effectively constrains policy exploration within a bounded vicinity of the optimized demonstrations. This synergy between rigorous physical refinement and guided exploration provides a robust foundation for achieving reliable grasping and naturally smooth manipulation trajectories within simulated environments.

\medskip
\noindent In summary, our contributions are as follows:
\begin{itemize}
\item We establish a complete and efficient pipeline for learning dexterous manipulation from monocular RGB videos, integrating 3D object asset acquisition, scale recovery, trajectory estimation, and policy learning, while demonstrating its robust applicability across unstructured environments.
\item We introduce a hierarchical refinement process that synergizes 2D and 3D tracking with geometric and Physical constraints. This two-stage approach significantly improves data quality and training efficiency, outperforming state-of-the-art baselines.
\item We demonstrate the effectiveness of our framework by achieving an average success rate of over 75\% across multiple synthetic manipulation tasks, while showing that the extracted motion priors can be successfully adapted to five heterogeneous dexterous hand embodiments through embodiment-aware retargeting and policy optimization.
\end{itemize}
\section{Related Works}
\label{sec:formatting}
\noindent \textbf{Dexterous Manipulation via Human Demonstrations} 
Learning dexterous manipulation skills from human demonstrations has long been a focal point in robotics as a means to circumvent the complexities of manual controller design. Early research primarily utilized high-fidelity demonstration data acquired via teleoperation or optical motion capture systems \citep{qin2023anyteleop,wang2024dexcap,taheri2020grab,chao2021dexycb,rakita2019shared,wang2013video,romero2022embodied,xie2023hmdo,zhan2024oakink2,yang2022oakink}. While these methods provide precise joint trajectories, the high hardware costs, specialized setup requirements, and time-consuming data collection processes significantly limit the scalability and diversity of the acquired skills. Consequently, there has been a growing interest in leveraging internet videos and egocentric datasets as more scalable data sources for human manipulation \citep{shaw2023videodex,8794127,qin2022dexmv,zhang2024dexgraspnet,10160982}. However, extracting reliable training samples from such unconstrained and uncurated data remains a formidable challenge. As highlighted by prior works \citep{on2025bigs,fan2024hold,zhang2021single}, independent estimation of hand and object poses \citep{wen2024foundationpose,yu2025dyn,tekin2019h+,lin2021end,potamias2025wilor} frequently results in significant spatial discrepancies and physical violations, such as interpenetration. This gap between raw vision-based estimation and physically plausible demonstration has prompted us to develop a two-stage optimization process, a direction we extend through our proposed refinement framework to ensure contact quality.

\medskip
\noindent \textbf{Reinforcement Learning with Human Priors} 
The high-dimensional action space of dexterous hands \citep{8794102,andrychowicz2020learning} makes reinforcement learning (RL) \citep{kaelbling1996reinforcement,schulman2017proximal,chen2022towards} from scratch computationally prohibitive. To accelerate exploration, mainstream approaches often integrate imitation learning with RL \citep{reddy2019sqil,rashidinejad2021bridging}, using human priors to guide the agent toward successful states. Among these, residual learning \citep{he2016deep,davchev2022residual,huang2024efficient,zhang2023efficient} has emerged as an effective paradigm. Instead of learning the entire control signal, residual methods focus on learning local refinements to a base motion or policy. This approach reduces the learning burden and allows the policy to handle the domain gap between human demonstrations and robotic execution, providing a robust framework for complex interaction tasks.

\medskip
\noindent \textbf{Video-based Skill Acquisition} 
The paradigm of robotic skill acquisition from human demonstrations has evolved rapidly \citep{shaw2023videodex,qin2022dexmv,mandikal2022dexvip,chen2025vividex,zhou2025you,chen2026dexterous,gavryushin2025maple}, yet the potential of video data as a direct source of physical expertise remains largely untapped. Many current Vision-Language-Action (VLA) models primarily leverage large-scale videos for pre-training visual-semantic alignments \citep{luo2025being,luo2026being}, failing to capture the authentic hand-object physical interaction information inherent in the footage. Ideally, video-derived demonstrations should achieve the fidelity and data weight of high-quality simulation or real-world samples; nevertheless, this remains challenging as prior frameworks are often confined to structured environments or static-camera settings to maintain tracking stability. Existing video-to-robot manipulation approaches primarily focus on trajectory reconstruction and imitation learning from human videos. Methods such as DexImit\citep{mu2026deximit} emphasize scalable robot trajectory synthesis and diffusion-based imitation, while DexMan\citep{hsieh2025dexman} focuses on RL-based trajectory tracking.

However, extending video-based skill acquisition to unstructured monocular settings remains fundamentally challenging, as monocular video priors are inherently noisy and physically inconsistent. Prior works have shown that reconstruction fidelity and policy performance degrade significantly under in-the-wild conditions involving camera motion and dynamic viewpoints, leading many existing approaches to rely on structured environments or static-camera assumptions for stable tracking and imitation. These challenges become particularly severe in egocentric or hand-held video settings, where motion blur, occlusion, and depth ambiguity often produce unstable contacts and physically infeasible interaction dynamics. In contrast, our work explicitly targets in-the-wild monocular videos, including egocentric camera motions, by directly distilling executable 3D hand-object interaction trajectories from raw RGB observations. Rather than relying solely on accurate trajectory imitation, we propose a physics-guided refinement and residual policy adaptation framework that bridges noisy visual priors with physically consistent manipulation behaviors. Through hierarchical geometric and kinetic refinement integrating spatial alignment, force-closure optimization, and residual reinforcement learning, our framework effectively narrows the gap between 2D visual observations and executable 3D dexterous manipulation.
\section{Method}
\begin{figure*}[t] 
    \centering
    \includegraphics[width=\textwidth]{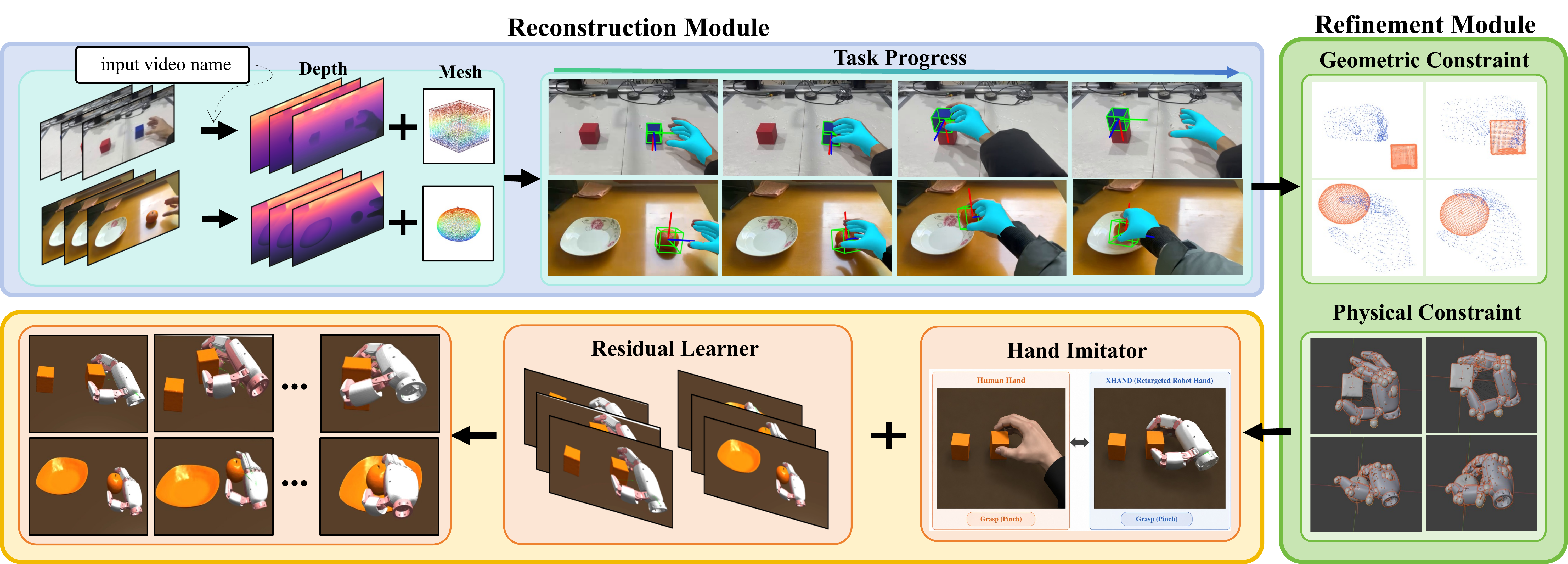} 
    \caption{\textbf{V2P-manip pipeline overview.} Our framework follows a hierarchical "perception-refinement-synthesis" workflow. The Reconstruction Module extracts temporal depth and mesh from monocular videos to perform joint hand-object pose tracking. These initial trajectories are subsequently optimized by the Refinement Module under geometric and physical constraints to ensure grasp plausibility. Finally, the Hand Imitator and Residual Learner jointly synthesize refined manipulation sequences to facilitate the efficient training of dexterous manipulation policies.}
    \label{fig:overview of V2P}
\end{figure*}
\subsection{Unified Perception and Object Reconstruction}
Given a monocular RGB sequence of human demonstrations $\{I_t\}_{t=1}^N$, we first employ a segmentor~\citep{kirillov2023segment} to extract precise binary masks, isolating target objects from the background. Subsequently, a depth estimator \citep{piccinelli2024unidepth,wang2026moge} infers per-pixel depth $\mathbf{D}_t$ and the corresponding camera intrinsic matrix $\mathbf{K}_t$ for each frame. To mitigate temporal inconsistencies inherent in per-frame intrinsic estimation, we establish a canonical intrinsic matrix by computing the global average, $\mathbf{\bar{K}} = \frac{1}{N} \sum_{i=1}^{N} \mathbf{K}_{i}$. The depth maps are then refined through a remapping operation, $D_t^{\text{cor}}(\mathbf{u}) = D_t(\mathbf{u}) \cdot \frac{\|\mathbf{\bar{K}}^{-1}\mathbf{u}\|_2}{\|\mathbf{K}_t^{-1}\mathbf{u}\|_2}$, which ensures spatial consistency across the sequence. Utilizing these rectified depth maps, object segments are back-projected into 3D space to recover their spatial coordinates $(x, y, z)$ via $x = (u - \bar{c}_x)d / \bar{f}_x$, $y = (v - \bar{c}_y)d / \bar{f}_y$, and $z = d$.
where $d$ denotes the corrected depth $D_t^{\text{cor}}(\mathbf{u})$, and $(\bar{f}_x, \bar{f}_y, \bar{c}_x, \bar{c}_y)$ represent the focal lengths and principal point from $\mathbf{\bar{K}}$. To ensure the fidelity of the reconstructed point clouds, spatial clustering is applied to filter out outliers, from which the absolute physical dimensions are estimated. Specifically, the length of the object's longest axis is defined as the target scale $s$. To recover the complete geometry, we employ SAM3D \citep{chen2025sam} to generate a coherent mesh from monocular views, which is then uniformly scaled to match $s$, yielding a metric-consistent digital twin $\mathbf{\mathcal{M}_{obj}}$. Notably, while the estimation of absolute scale benefits from depth accuracy, our pipeline does not strictly depend on perfect metric precision. In cases of depth bias, the system remains capable of tracking trajectories within a scaled virtual space, effectively preserving the internal motion structure and physical feasibility.
\subsection{Pose estimation and Optimization}
To recover coordinated hand-object motion, we conduct vision-based tracking and subsequently introduce a dual-stage refinement layer to enforce both geometric alignment and Physical stability.

\noindent \textbf{Object-Hand Tracking.}
Based on the data obtained from the preparation phase, we leverage FoundationPose \citep{wen2024foundationpose} to estimate the 6D object pose, defined as $T_t \in \mathbb{SE}(3)$. However, under severe occlusions or rapid motions, FoundationPose often suffers from tracking drift and abrupt pose jumps due to unreliable internal scoring metrics. To mitigate these tracking failures, our system incorporates SpatialTracker \citep{xiao2024spatialtracker} to extract robust, pixel-level temporal correspondences, which regularize and stabilize the trajectory estimation, where a more comprehensive architectural description can be found in Appendix~\ref{subsec:object_pose}.

To capture the hand's kinematic state, we employ a robust pose tracker \citep{yu2025dyn} to estimate the parameters of the MANO \citep{romero2022embodied} model. For each frame $t$, the tracker regresses the shape parameters $\beta \in \mathbb{R}^{10}$ and the pose parameters $\theta_t \in \mathbb{R}^{15 \times 3}$, the latter of which represent the relative rotations of the hand joints. The reconstruction process yields a temporal hand trajectory $\mathcal{T}^h = \{ \tau_t^h \}_{t=0}^T$, where each state $\tau_t^h$ is formulated as:
\begin{equation}
\tau_t^h = (\theta_t, \beta, \mathbf{R}_t^h, \mathbf{t}_t^h)
\end{equation}
\noindent Here, $\mathbf{R}_t^h \in SO(3)$ and $\mathbf{t}_t^h \in \mathbb{R}^3$ denote the global orientation and translation of the wrist, respectively. This sequence of parameterized hand meshes serves as a robust kinematic prior, providing the geometric foundation for the following stages.

\noindent \textbf{Geometric Constraint.} Based on the initial hand and object pose estimations, we implement a hierarchical, dual-stage geometric optimization framework to resolve monocular spatial misalignments. The first stage rectifies systemic biases by optimizing a unified scale factor and rigid offsets across all contact frames, incorporating an anatomically weighted contact loss that prioritizes key grasping sites (e.g., thumb and index finger) while preserving the original relative motion structure. Building upon this, the second stage performs fine-grained, per-frame local refinement of pose and translation parameters. To ensure precise mesh-to-surface alignment with high tracking fidelity, the objective function jointly optimizes the localized contact constraints, temporal smoothness terms, and a regularization term penalizing excessive deviations from the raw estimations, with further elaboration provided in Appendix~\ref{subsec:constraints_hypers}.

\noindent \textbf{Physical Constraint.} To bridge the morphological gap between human demonstrations and robotic embodiments, we implement a cross-embodiment motion retargeting pipeline. 

\noindent We formulate the motion retargeting task as a non-linear optimization problem \citep{qin2023anyteleop}. For each frame $t$, the system solves for the optimal joint configuration $\mathbf{q}$ by minimizing the weighted distance between the robot's joints and tips and their corresponding targets derived from the MANO model:
\begin{equation}
\min_{\mathbf{q}} \sum_{i \in \text{joints}} w_i \| \text{FK}(\mathbf{q})_i - P_{\text{target},i} \|_2
\end{equation}
where $\mathbf{q}$ denotes the joint angles of the dexterous hand, and $\text{FK}(\cdot)$ represents the forward kinematics chain. We assign higher weights $w_i$ to prioritize functional digits (e.g., thumb and index) to preserve the original manipulation intent.

\noindent To ensure physical feasibility, we implement an adaptive trajectory refinement pipeline within the Isaac Gym simulation environment. We first perform per-frame stability validation, directly retaining trajectory frames where the object's 6D pose remains invariant within a predefined tolerance over fixed simulation steps. For frames exhibiting instability, we introduce a physics-grounded optimization process within the Task Wrench Space (TWS) \citep{curobo_report23,chen2024bodex} to restore physical feasibility. Specifically, the framework samples random external force directions and utilizes the GJK algorithm \citep{gilbert1988fast} to query proximal contact locations and local surface normals. A joint objective function is then formulated to minimize grasp energy, encouraging the synthesized contact wrenches within the local friction cones to counteract external disturbances. Concurrently, the objective enforces a critical distance threshold via GJK alignment to maintain contact proximity and prevent mesh penetration.This optimization improves physical plausibility while maintaining kinematic consistency for contact-rich manipulation. Detailed formulations and hyperparameter settings are provided in Appendix~\ref{subsec:constraints_hypers}.

\subsection{Residual Reinforcement Learning}
To bridge the gap between structured kinematic priors and physical execution, we formulate the dexterous manipulation task as a Markov Decision Process (MDP), where the state space $s_t$ integrates robot proprioception, object geometric features, and goal-driven task poses. The control action $a_t$ is decomposed into a reference base motion and a learned adaptive residual, defined as:
\begin{equation}
    a_t = a_{\text{base},t} + \pi_{\phi}(s_t)
\end{equation}
where the base action $a_{\text{base},t}$ is provided by a pre-trained hand imitator. This imitator is trained on the foundational data sources of ManipTrans \citep{li2025maniptrans}, augmented with synthesized dataset to establish a high-fidelity motion prior. The residual policy $\pi_{\phi}(s_t)$ subsequently learns to compensate for unmodeled dynamics and domain gaps.

\noindent We employ Proximal Policy Optimization (PPO) \citep{schulman2017proximal} to train the networks within the Isaac Gym simulator \citep{makoviychuk2021isaac}, leveraging highly parallelized simulation environments to accelerate policy convergence. To ensure training stability, we implement an early termination mechanism \citep{peng2021amp} and a curriculum learning scheme \citep{bengio2009curriculum}. The overall reward function $R_t$ is formulated as a composite of imitation and contact-driven terms:
\begin{equation}
    R_t = r_{\text{imit}} + \mathbb{I}_{\text{res}} \cdot r_{\text{cont}}
\end{equation}
where the indicator $\mathbb{I}_{\text{res}} \in \{0, 1\}$ activates exclusively during residual refinement. The imitation term $r_{\text{imit}}$ enforces reference trajectory tracking via a weighted sum of exponential alignment kernels:
\begin{equation}
    r_{\text{imit}} = \sum_{k \in \mathcal{K}} w_k \exp\left( -\alpha_k \|x_k - \hat{x}_k\|_{\star} \right)
\end{equation}
where $w_k$ and $\alpha_k$ are the weight and scale for component $k$. The distance metric $\|\cdot\|_{\star}$ instantiates as the $L_2$-norm for positions, $L_1$-norm for velocities, or absolute geodesic angle error for orientations. 

Specifically, the tracked component set $\mathcal{K}$ adapts to the training phase: for base model training ($\mathbb{I}_{\text{res}} = 0$), $\mathcal{K}_{\text{base}} = \{w, f, j\}$ focuses solely on the wrist ($w$), fingertips ($f$), and joints ($j$) for motion synthesis. In the residual phase ($\mathbb{I}_{\text{res}} = 1$), it expands to $\mathcal{K}_{\text{res}} = \{w, f, j, o\}$ to incorporate object pose tracking ($o$). Concurrently, $r_{\text{cont}}$ is introduced to regularize physical interaction and contact force:
\begin{equation}
    r_{\text{cont}} = \exp \left( -\frac{\alpha_c}{\left\| \sum_{i \in \text{tips}} \omega_i F_{\text{net},i} \right\| + \epsilon} \right)
\end{equation}
where $F_{\text{net},i}$ represents the net contact force and $\omega_i$ denotes proximity-based weights. This formulation ensures that the base model provides a kinematic foundation, while the residual policy refines it into a dynamically stable grasp through active object interaction. To bolster robustness, we apply Domain Randomization (DR) over physical parameters including gravity, friction, and object mass.
\section{Experiments}
\subsection{Object Pose Estimation}
We evaluate our pipeline’s performance through two primary lenses: 1) standard 6D trajectory recovery from raw monocular egocentric videos in the TACO dataset \citep{liu2024taco} to establish reliable digital twins; 2) robustness stress-testing under adversarial conditions including heavy occlusions, cluttered backgrounds, and motion blur.

\noindent \textbf{Qualitative Analysis on Challenging Scenarios.} 
As illustrated in appendix, our method maintains superior tracking stability over the FoundationPose baseline under various adversarial conditions. While FoundationPose frequently suffers from track loss during heavy hand-object occlusions, our multi-hypothesis strategy leverages temporal motion priors to ensure trajectory continuity. Moreover, by incorporating pixel-level correspondences from SpatialTracker, our pipeline effectively mitigates environmental noise and abrupt motion blur in scenarios.

\noindent \textbf{Quantitative Evaluation on Egocentric Videos.}
We benchmark our performance against FoundationPose and SpatialTracker on the TACO dataset using standard metrics:  Chamfer Distance (CD), ADD-S \citep{xiang2017posecnn}, Failure Rate (FR), Relative Translation Error (RTE), Relative Rotation Error (RRE), and a unified Stability Index (SI)\citep{zeng20173dmatch}. The RTE, RRE, and SI quantify trajectory jitter and continuity between consecutive frames. As presented in Table \ref{table:taco_results}, our method achieves the highest precision and robustness, achieving an ADD-S of 80.52\% and a lower Failure Rate of 7.27\%. Furthermore, our approach yields the most favorable SI (0.535), indicating superior temporal consistency across these challenging sequences. 
\begin{table}[h]
\centering
\caption{Performance Comparison on the TACO Dataset}
\label{table:taco_results}
\resizebox{\columnwidth}{!}{
\begin{tabular}{lcccccc}
\toprule
& \multicolumn{2}{c}{\textbf{Pose Accuracy}} & \textbf{Robustness} & \multicolumn{3}{c}{\textbf{Temporal Smoothness}} \\
\cmidrule(lr){2-3} \cmidrule(lr){4-4} \cmidrule(lr){5-7}
Method & CD (cm) $\downarrow$ & ADD-S (\%) $\uparrow$ & FR (\%) $\downarrow$ & RTE (cm) $\downarrow$ & RRE (rad) $\downarrow$ & SI $\uparrow$ \\
\midrule
FoundationPose & 1.685 & 76.98 & 10.52 & 0.926 & \textbf{0.052} & 0.529 \\
SpatialTracker  & 1.829 & 63.65 & 28.45 & \textbf{0.765} & 0.085          & 0.532 \\
\textbf{Ours}   & \textbf{1.447} & \textbf{80.52} & \textbf{7.27}  & 0.897          & 0.053          & \textbf{0.535} \\
\bottomrule
\end{tabular}
}
\end{table}

\subsection{Policy Learning Evaluation}

\begin{figure}[htbp]
    \centering
    \includegraphics[width=0.8\columnwidth]{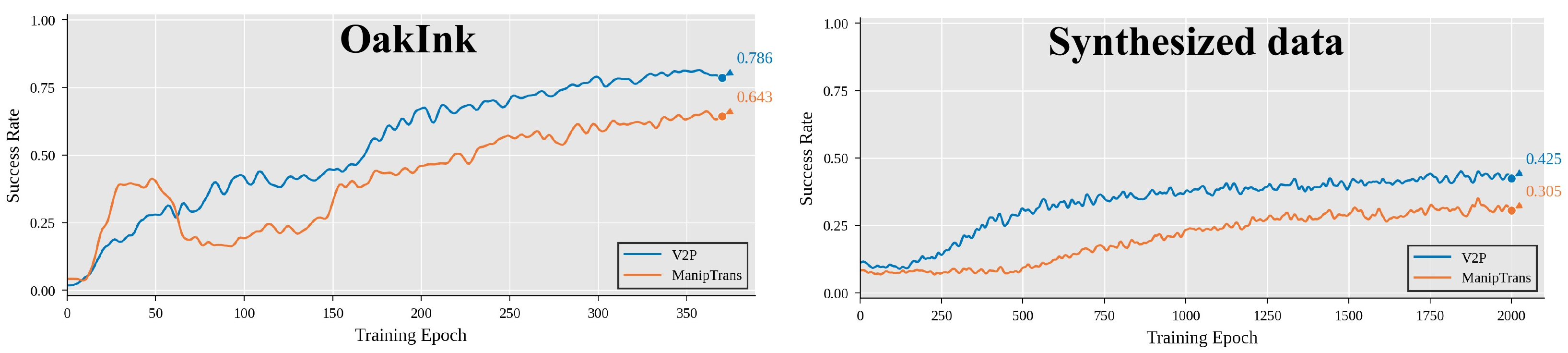}
    \caption{Policy learning curves and performance comparison. Ours exhibits significantly faster convergence and higher success rates compared to ManipTrans on both synthesized and OakInk datasets.}
    \label{fig:performance_comparison}
\end{figure}

We evaluate the efficacy of our framework on the OakInk \citep{zhan2024oakink2} benchmark, incorporating augmented initialization, reset exploration, and synthesized data to boost learning performance \citep{peng2018deepmimic}. As illustrated in Fig.\ref{fig:performance_comparison}, our method demonstrates superior convergence speed and achieves higher peak success rates compared to the ManipTrans \citep{li2025maniptrans} baseline in both synthesized and transfer scenarios. 
\begin{table}[htbp]
\centering
\caption{Quantitative Evaluation on the OakInk Benchmark. $E_{R}$ and $E_{T}$ denote relative rotation and translation errors, where subscripts $r$ and $l$ represent the right and left hands, respectively.}
\label{table:oakink_eval}
\small 
\setlength{\tabcolsep}{8pt} 
\begin{tabular}{lccccc}
\toprule
Method & SR $\uparrow$ & $E_{rR}$ (rad) $\downarrow$ & $E_{rT}$ (m) $\downarrow$ & $E_{lR}$ (rad) $\downarrow$ & $E_{lT}$ (m) $\downarrow$ \\
\midrule
ManipTrans & 0.417 & 0.197 & 0.00869 & \textbf{0.277} & 0.0139 \\
\textbf{Ours} & \textbf{0.483} & \textbf{0.182} & \textbf{0.00647} & 0.283 & \textbf{0.0117} \\
\bottomrule
\end{tabular}
\end{table}

\noindent The quantitative results summarized in Table \ref{table:oakink_eval} further confirm that our approach consistently outperforms the baseline across most key metrics. Notably, our method achieves a significant boost in the overall Success Rate (SR) while maintaining high precision in relative rotation ($E_{R}$) and translation ($E_{T}$). These results validate that the integration of high-quality trajectory priors and structured exploration effectively facilitates policy learning in complex, contact-rich dexterous manipulation tasks.

\subsection{Ablation Study}

We conduct an ablation study using the XHand across eight diverse synthesized manipulation tasks to evaluate the individual contributions of our core components: the Geometric Constraint ($C_1$) and the Physical Constraint ($C_2$). Specifically, w/o $C_1$ denotes the removal of global pose alignment and hand-object proximity priors, while w/o $C_2$ represents the exclusion of fine-grained contact refinement and force-based optimization. 

\noindent As shown in Table \ref{tab:multi_task_results}, the baseline models lacking these constraints exhibit significant performance degradation. The results indicate that $C_1$ establishes the necessary spatial foundation for hand-object coordination, without which most tasks fail to initialize. Furthermore, the inclusion of $C_2$ is essential for refining complex interactions, particularly in high-precision scenarios. The full V2P framework consistently achieves the highest average success rate (79.3\%) on XHand, validating that the synergy between geometric alignment and physical optimization is key to robust dexterous manipulation.

\begin{table*}[!ht]
  \centering
  \caption{Quantitative multi-task evaluation and ablation results on XHand. Success rates are reported as successful trials out of 100 total episodes. $C_1$ and $C_2$ refer to Geometric and Physical constraints.}
  \label{tab:multi_task_results}
  \small 
  \begin{tabular*}{\textwidth}{l@{\extracolsep{\fill}}cccc}
    \toprule
    \textbf{Task Name} & \textbf{Origin} & \textbf{w/o $C_1$} & \textbf{w/o $C_2$} & \textbf{V2P (Ours)} \\
    \midrule
    Stacking Blocks      & 0 / 100 (0.0\%) & 0 / 100 (0.0\%) & 68 / 100 (68.0\%) & \textbf{99 / 100 (99.0\%)} \\
    Pouring water (cup)  & 0 / 100 (0.0\%) & 0 / 100 (0.0\%) & 62 / 100 (62.0\%) & \textbf{92 / 100 (92.0\%)} \\
    Placing orange         & 0 / 100 (0.0\%) & 0 / 100 (0.0\%) & 100 / 100 (100.0\%) & \textbf{100 / 100 (100.0\%)} \\
    Pouring water (mug)  & 0 / 100 (0.0\%) & 0 / 100 (0.0\%) & 65 / 100 (65.0\%) & \textbf{85 / 100 (85.0\%)} \\
    Using brush          & 0 / 100 (0.0\%) & 0 / 100 (0.0\%) & 74 / 100 (74.0\%) & \textbf{94 / 100 (94.0\%)} \\
    Hammering            & 0 / 100 (0.0\%) & 0 / 100 (0.0\%) & 7 / 100 (7.0\%)  & \textbf{51 / 100 (51.0\%)} \\
    Using Toothbrush     & 0 / 100 (0.0\%) & 0 / 100 (0.0\%) & 14 / 100 (14.0\%) & \textbf{33 / 100 (33.0\%)} \\
    Holding Microphone   & 0 / 100 (0.0\%) & 0 / 100 (0.0\%) & 45 / 100 (45.0\%) & \textbf{80 / 100 (80.0\%)} \\
    \midrule
    \textbf{Avg. Success Rate} & 0.0\% & 0.0\% & 54.4\% & \textbf{79.3\%} \\ 
    \bottomrule
  \end{tabular*} 
\end{table*}

\subsection{Generalization and Visualization Results}
\begin{figure*}[t] 
    \centering
    \includegraphics[width=\textwidth]{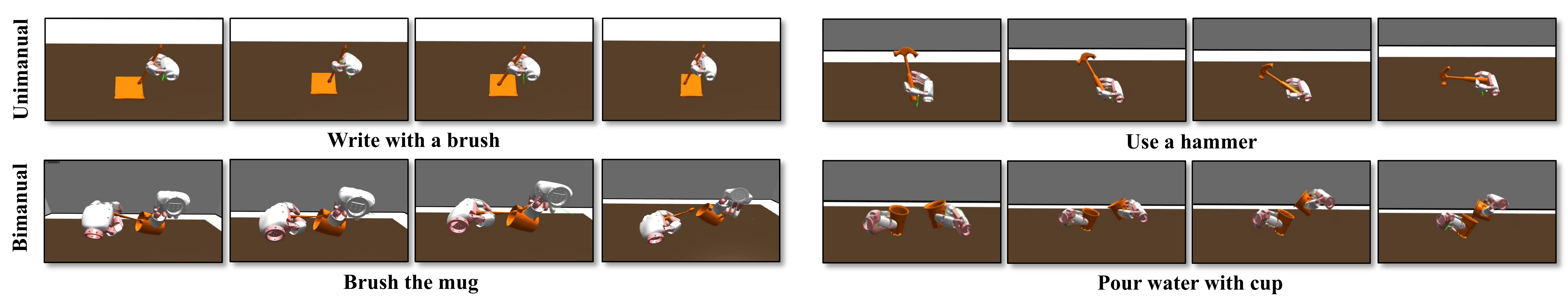} 
    \caption{\textbf{Qualitative results of synthesized dexterous manipulation.} We showcase recovered trajectories for unimanual (top) and bimanual (bottom) tasks from monocular videos, where our method produces physically-plausible results across various complex scenarios.}
    \label{fig:Visual_results_1}
\end{figure*}
We further evaluate the adaptability of our framework across four additional robotic platforms, including Allegro, Arti-MANO, Inspire, and Shadow hands. This multi-embodiment evaluation assesses whether the motion priors extracted by V2P can be effectively retargeted and optimized under diverse kinematic structures and hand morphologies.
\begin{table*}[!ht]
  \centering
  \caption{Quantitative results of multi-embodiment adaptation. Success rates are reported as the number of successful completions out of 100 total episodes for each robotic platform.}
  \label{tab:success_counts}
  \scriptsize 
  \begin{tabular*}{\textwidth}{l@{\extracolsep{\fill}}ccccc}
    \toprule
    \textbf{Task} & \textbf{Allegro} & \textbf{Arti-MANO} & \textbf{Inspire} & \textbf{Shadow} & \textbf{Mean SR} \\ 
    \midrule
    Stacking Blocks      & 48 / 100 (48.0\%) & 62 / 100 (62.0\%) & 98 / 100 (98.0\%) & 74 / 100 (74.0\%) & 70.5\% \\
    Placing Orange         & 66 / 100 (66.0\%) & 98 / 100 (98.0\%) & 88 / 100 (88.0\%) & 96 / 100 (96.0\%) & 87.0\% \\
    Pouring with Mug     & 70 / 100 (70.0\%) & 86 / 100 (86.0\%) & 66 / 100 (66.0\%) & 44 / 100 (44.0\%) & 66.5\% \\
    Using Brush          & 40 / 100 (40.0\%) & 96 / 100 (96.0\%) & 56 / 100 (56.0\%) & 18 / 100 (18.0\%) & 52.5\% \\
    Holding Microphone   & 36 / 100 (36.0\%) & 70 / 100 (70.0\%) & 86 / 100 (86.0\%) & 76 / 100 (76.0\%) & 67.0\% \\
    \midrule
    \textbf{Avg. Success Rate} & \textbf{52.0\%} & \textbf{82.4\%} & \textbf{78.8\%} & \textbf{61.6\%} & \textbf{68.7\%} \\ 
    \bottomrule
  \end{tabular*}
\end{table*}
\noindent The performance variance observed in Table \ref{tab:success_counts} stems from fundamental differences in kinematics, link dimensions, and fingertip geometries among the evaluated platforms. The Arti-MANO achieves the highest overall success rate (82.4\%) largely due to its anthropomorphic design; its link proportions and joint hierarchies closely align with the human MANO model used in our trajectory optimization, facilitating a natural mapping of dexterous poses in contact-rich tasks. In contrast, the high-dimensional configurations of the Shadow and Allegro hands introduce significant complexity, as their increased degrees of freedom expand the action space dimensionality and make policy convergence more challenging. Furthermore, the Allegro Hand’s bulky physical scale and thicker fingertips present significant geometric constraints, often leading to self-collisions or mesh penetrations when handling fine-grained objects. In our physics-based simulation, such geometric interference frequently triggers early termination to prevent unphysical states, accounting for the lower success rates in high-precision manipulation tasks.
\section{Limitations}
While our framework successfully facilitates dexterous manipulation learning from monocular videos, certain constraints remain. Currently, our methodology is tailored for rigid-body objects; extending it to articulated or deformable entities remains unexplored due to their high degrees of freedom and complex contact physics. Furthermore, the simulation enforces a floating-base assumption that omits specific robot arm morphologies, requiring external inverse kinematics solvers to manage joint reachability. Directly incorporating full-body kinematic and dynamic constraints into the policy learning pipeline represents a vital next step to ensure end-to-end trajectory feasibility within the optimization loop.
\section{Discussion}
V2P-Manip provides a unified pipeline for converting unconstrained monocular human videos into physically feasible dexterous manipulation trajectories. By integrating reconstruction and refinement modules, our approach eliminates the need for specialized tracking hardware or motion-capture suites, thereby reducing the gap between raw RGB perception and trajectory-based policy learning. Experimental results demonstrate that V2P-Manip facilitates complex skill acquisition while maintaining natural motion characteristics across heterogeneous robotic hand embodiments. Furthermore, by enforcing physically grounded consistency during trajectory reconstruction and optimization, the proposed framework improves robustness under noisy and partial observations, while leveraging automated trajectory augmentation to compensate for inherent kinematic discrepancies. Future work will focus on scaling this framework to full arm-hand robot embodiments and extending it to articulated and deformable objects. By incorporating diverse object geometries and robot morphologies, we seek to systematically exploit the latent physical constraints within human demonstrations to enhance the scalability of dexterous embodied intelligence.
\clearpage


\bibliography{example}  
\clearpage
\appendix 


\counterwithin{figure}{section}
\counterwithin{table}{section}
\counterwithin{equation}{section}

\section*{Supplementary Material for V2P-Manip}
\label{sec:supplementary_title}

To provide further comprehensive analysis, this supplementary material expands upon the methodology and experimental results of the main text. An overview of the contents is structured as follows:

\vspace{0.5em}
\hrule
\vspace{0.8em}
\noindent\textbf{Supplementary Overview}
\begin{itemize}
    \item \textbf{Section~\ref{sec:method_details}: Method Details}
    \begin{itemize}
        \item Section~\ref{subsec:depth_reconstruction}: Depth Estimation and Size Reconstruction
        \item Section~\ref{subsec:object_pose}: Object Pose Estimation
        \item Section~\ref{subsec:vggt_extrinsics}: VGGT Extrinsic Estimation
        \item Section~\ref{subsec:taco_metrics}: TACO Evaluation Metrics
        \item Section~\ref{subsec:constraints_hypers}: Geometric \& Physical Constraints
        \item Section~\ref{subsec:rl_training}: Reinforcement Learning Training
    \end{itemize}
    
    \item \textbf{Section~\ref{sec:visual_results}: Visual Results}
    \begin{itemize}
        \item Section~\ref{subsec:cross_embodiment}: Cross-Embodiment Results
        \item Section~\ref{subsec:egocentric_results}: Egocentric Results
        \item Section~\ref{subsec:real_robot_execution}: Real-Robot Trajectory Execution
    \end{itemize}
    
    \item \textbf{Section~\ref{sec:limitations}: Additional Analysis}
\end{itemize}
\hrule
\vspace{1.5em}


\section{Method Details}
\label{sec:method_details}

\subsection{Depth Estimation and Size Reconstruction}
\label{subsec:depth_reconstruction}

To lift 2D video frames into aligned 3D meshes, we evaluate state-of-the-art monocular depth estimators (UniDepth vs. MeGo2) against the ground-truth object geometry from the TACO dataset. As demonstrated by the quantitative results in Table~\ref{tab:depth_errors}, MeGo2 delivers a significantly lower mean size error, achieving a superior balance between depth accuracy and computational efficiency for hand-object interaction scenes. Consequently, we adopt MeGo2 as our primary depth estimator to anchor the spatial scale and perform robust 3D size reconstruction across sequential video frames.

\begin{table}[h]
\centering
\caption{Quantitative comparison of depth estimators on spatial scale reconstruction.}
\label{tab:depth_errors}
\begin{tabular}{lc}
\toprule
\textbf{Method} & \textbf{Mean Size Error (\%)} $\downarrow$ \\
\midrule
UniDepth        & 0.0682 \\
MeGo2           & \textbf{0.0488} \\
\bottomrule
\end{tabular}
\end{table}

\subsection{Object Pose Estimation}
\label{subsec:object_pose}
Based on the tracked trajectories estimated by FoundationPose and SpatialTracker, we introduce a rigid motion consistency check to filter out noisy points, thereby yielding a reliable rigid feature subset $\mathcal{P}_{r}$. To accurately capture inter-frame object dynamics, the optimal rigid transformation $\Delta T$ is solved by minimizing the Procrustes distance:
\begin{equation}
\Delta T^* = \arg\min_{\Delta T} \sum_{i \in \mathcal{S}} \| \Delta T \cdot p_i^t - p_i^{t+1} \|^2
\end{equation}
where $\mathcal{S}$ denotes a point cloud sampling strategy set consisting of the full set $\mathcal{P}_{r}$ and $K$ randomly sampled subsets $\mathcal{P}_{\text{sub}, k} \, (k=1, \dots, K)$. This parallel computation generates motion candidates, forming the motion-prior candidate set $\mathcal{T}_{m,t} = \{\Delta T_m \cdot T_{t-1}\}_{m=1}^{K+1}$.

\noindent Subsequently, the system constructs a hybrid candidate pose space $\mathcal{T}_t$ by integrating motion priors with current visual observations:
\begin{equation}
\mathcal{T}_t = \{ T_{t-1} \} \cup \mathcal{T}_{m,t} \cup \mathbbm{1}_q \{ T_{r,k} \}_{k=1}^{K+2}
\end{equation}
where $\{T_{r,k}\}$ denotes the set of refined pose candidates generated by FoundationPose based on the current frame. The indicator function $\mathbbm{1}_q$ filters out these refinement candidates in low-quality frames.

\noindent To automate the evaluation of frame quality, we implement an iterative drift-detection loop. At time step $t$, let $\tau_t$ be the 2D keypoint trajectory from SpaTracker and $\hat{\tau}_t$ be the re-projected trajectory derived from our hypothesis model. We define the tracking residual as $\mathcal{R}_t = \| \tau_t - \hat{\tau}_t \|_2$. A frame is categorized as low-quality (i.e., $\mathbbm{1}_q = 0$) if $\mathcal{R}_t > \delta$, where $\delta$ is a predefined threshold. The overall sequence reliability is quantified by the ratio $\eta = \frac{1}{N} \sum_{t=1}^N (1 - \mathbbm{1}_{q,t})$.

\noindent Finally, the optimal pose $T_t^*$ is selected from the candidate space $\mathcal{T}_t$ by maximizing the confidence score provided by the FoundationPose scoring network:
\begin{equation}
T_t^* = \arg\max_{T \in \mathcal{T}_t} \text{Score}(T | I_t, M_t)
\end{equation}

\noindent This multi-hypothesis strategy, coupled with occlusion-aware switching, effectively compensates for the failure of single-view observations during dense interactions, ensuring both temporal continuity and high precision.
\begin{figure}[htbp]
    \centering
    \includegraphics[width=\columnwidth]{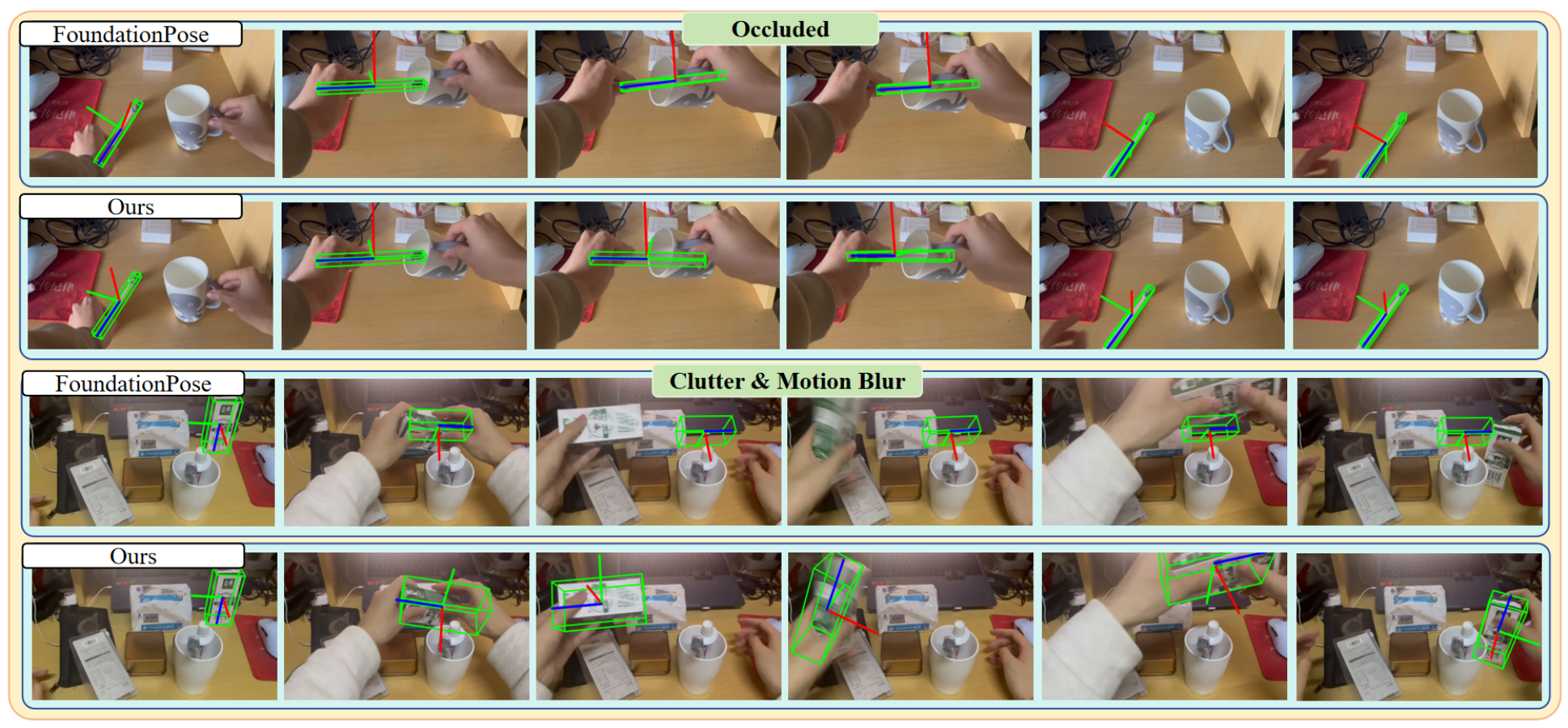} 
    \caption{Qualitative comparison. Our method maintains robust tracking under occlusion, clutter, and motion blur compared to FoundationPose.}
    \label{fig:qualitative_analysis}
\end{figure}

\begin{figure}[htbp]
    \centering
    \includegraphics[width=\columnwidth]{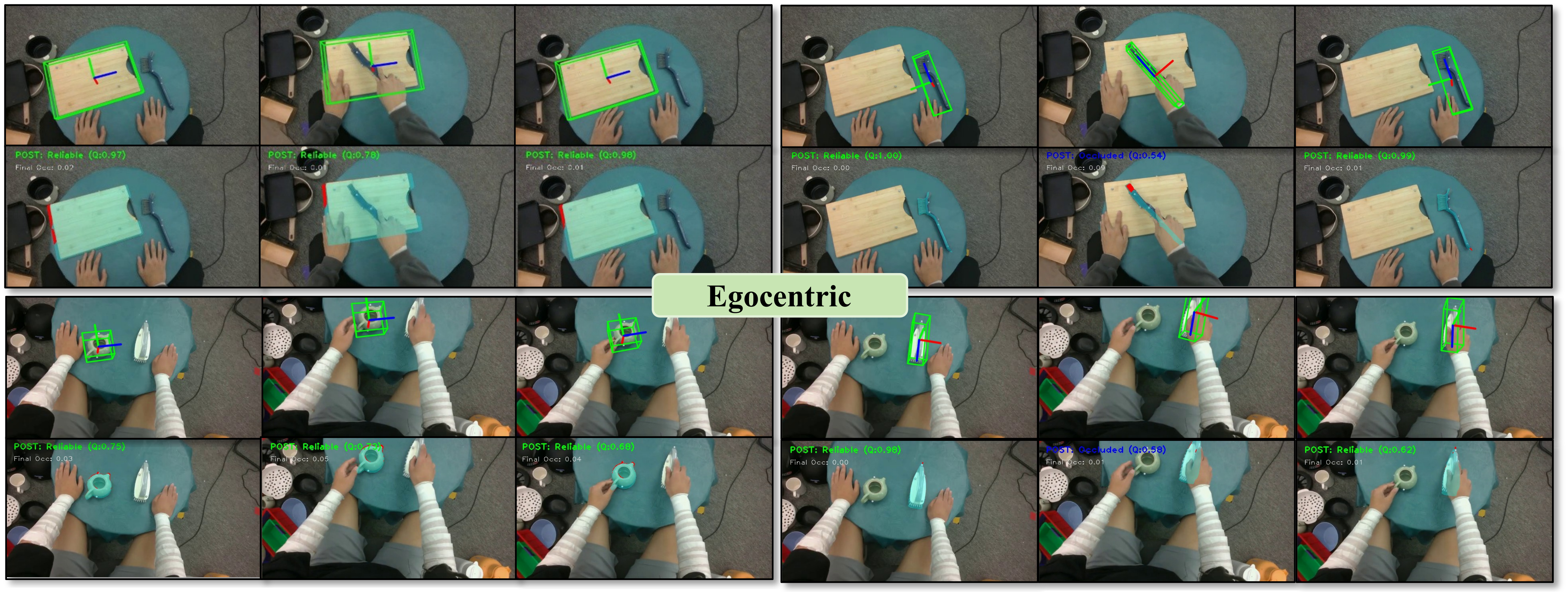}
    \caption{\textbf{Qualitative results of egocentric object pose tracking.} 
    We illustrate our pose tracking performance across sequential frames. 
    Odd rows display the 2D Oriented Bounding Boxes (OBB) indicating object localization. 
    Even rows show the reconstructed object meshes projected onto the image plane using our estimated 3D rigid transformations, demonstrating robustness under hand-object occlusions.}
    \label{fig:Egocentric}
\end{figure}

\subsection{VGGT Extrinsic Estimation}
\label{subsec:vggt_extrinsics}

Since VGGT processes video sequences in a chunk-wise manner, directly applying it to long-horizon videos often exceeds GPU memory limits. To address this limitation, we partition the input video into 30-frame chunks with a 10-frame temporal overlap $\mathcal{O}_k$ between consecutive segments. Since the extrinsic parameters estimated by VGGT are initialized as identity transformations ($I \in \mathrm{SE}(3)$) at the first frame of each chunk, independently estimated local trajectories exhibit discontinuities across chunk boundaries.

Formally, let $T_{k-1,t}, T_{k,t} \in \mathrm{SE}(3)$ denote the estimated camera extrinsic matrices at frame $t$ within chunks $k-1$ and $k$, respectively. We estimate the optimal relative alignment $\Delta T_k \in \mathrm{SE}(3)$ by minimizing the transformation discrepancy over the overlap region:
\begin{equation}
\min_{\Delta T_k}
\sum_{t \in \mathcal{O}_k}
\left|
\log \left(
\Delta T_k T_{k,t} T_{k-1,t}^{-1}
\right)
\right|_2^2
\label{eq:extrinsic_alignment}
\end{equation}
where $\log(\cdot)$ denotes the logarithm map from $\mathrm{SE}(3)$ to its Lie algebra $\mathfrak{se}(3)$. By iteratively propagating $\Delta T_k$ across sequential chunks, we obtain a globally consistent and temporally smooth camera extrinsic trajectory for the entire video.

\subsection{TACO Evaluation Metrics}
\label{subsec:taco_metrics}

To comprehensively evaluate the accuracy and temporal consistency of the estimated 6D object poses against the ground-truth trajectories from the TACO dataset, we employ a streamlined suite of geometric and temporal metrics. Let $M$ denote the total number of frames in the sequence. For the $i$-th frame, let $T_i, \hat{T}_i \in \mathbb{SE}(3)$ represent the ground-truth and estimated object poses, respectively, and let $t_i, \hat{t}_i \in \mathbb{R}^3$ be their corresponding 3D translation vectors. We denote the ground-truth and estimated trajectory point sets as $\mathcal{P} = \{t_i\}_{i=1}^M$ and $\hat{\mathcal{P}} = \{\hat{t}_j\}_{j=1}^M$. Furthermore, let $\mathcal{M}$ represent the downsampled 3D object mesh containing $N$ vertices.

\noindent\textbf{Trajectory Chamfer Distance (CD).} To decouple pure global translation errors from the mesh geometry, we calculate the bidirectional Chamfer Distance directly between the ground-truth and estimated 3D trajectory position sets:
\begin{equation}
e_{\text{CD}} = \frac{1}{2} \left( \frac{1}{M}\sum_{t \in \mathcal{P}} \min_{\hat{t} \in \hat{\mathcal{P}}} \|t - \hat{t}\|_2 + \frac{1}{M}\sum_{\hat{t} \in \hat{\mathcal{P}}} \min_{t \in \mathcal{P}} \|\hat{t} - t\|_2 \right)
\label{eq:traj_cd}
\end{equation}

\noindent\textbf{Average Distance with Symmetry (ADD-S).} To evaluate the localized 6D pose accuracy per frame, we employ the Average Distance with Symmetry (ADD-S). For the $k$-th frame, let $R_k, t_k$ and $\hat{R}_k, \hat{t}_k$ denote the rotation and translation components of the ground-truth pose $T_k$ and estimated pose $\hat{T}_k$, respectively. The frame-level ADD-S error is defined as:
\begin{equation}
e_{ADD-S} = \frac{1}{N} \sum_{p \in \mathcal{M}} \min_{q \in \mathcal{M}} \| (R_k p + t_k) - (\hat{R}_k q + \hat{t}_k) \|_2
\label{eq:adds}
\end{equation}
where $\mathcal{M}$ represents the downsampled 3D object mesh containing $N$ vertices.

To aggregate performance across the full sequence of $M$ frames, we report the \textbf{AUC} (Area Under the Accuracy-Threshold Curve), expressed as a percentage. The final AUC score is computed by integrating the fraction of accurate frames over the threshold interval $\tau \in [0, \tau_{\max}]$, where $\tau_{\max} = 10\text{~cm}$:
\begin{equation}
\text{AUC}_{\text{ADD-S}} = \frac{1}{\tau_{\max}} \int_{0}^{\tau_{\max}} \left( \frac{1}{M} \sum_{k=1}^M \mathbb{I}(e_k < \tau) \right) d\tau
\label{eq:auc_combined}
\end{equation}
where $\mathbb{I}(\cdot)$ represents the indicator function that outputs $1$ if the condition is satisfied and $0$ otherwise.

\noindent\textbf{Failure Rate (FR).} A pose estimation is classified as a tracking failure if its ADD-S error exceeds a predefined clearing threshold of 5~cm ($e_{\text{ADD-S}} > 5$~cm). FR represents the percentage of failed frames across the sequence.

\noindent\textbf{Stability Index (SI).} To evaluate tracking smoothness and quantify high-frequency jitter, we introduce a sequence-level Stability Index (SI). For the $k$-th frame within a specific sequence $j$, the relative motion discrepancy between consecutive frames is defined as $\Delta T_{\text{err}, k}^j = (\Delta T_k^j)^{-1} \Delta \hat{T}_k^j$, where $\Delta T_k^j = (T_k^j)^{-1} T_{k+1}^j$ and $\Delta \hat{T}_k^j = (\hat{T}_k^j)^{-1} \hat{T}_{k+1}^j$ represent the ground-truth and estimated inter-frame relative transformations, respectively. 

From $\Delta T_{\text{err}, k}^j$, we extract the frame-level Relative Translation Error ($e_{\text{RTE}, k}^j$) and Relative Rotation Error ($e_{\text{RRE}, k}^j$). For a sequence $j$ with $M_j$ frames, the sequence-averaged errors are computed as $\overline{e}_{\text{RTE}}^j = \frac{1}{M_j-1} \sum_{k=1}^{M_j-1} e_{\text{RTE}, k}^j$ and $\overline{e}_{\text{RRE}}^j = \frac{1}{M_j-1} \sum_{k=1}^{M_j-1} e_{\text{RRE}, k}^j$. The stability score for this specific sequence is then defined as:
\begin{equation}
\text{SI}_j = \frac{1}{2} \left( \exp\left( - \frac{\overline{e}_{\text{RTE}}^j}{\sigma_{\text{trans}}} \right) + \exp\left( - \frac{\overline{e}_{\text{RRE}}^j}{\sigma_{\text{rot}}} \right) \right)
\label{eq:si_metric_sequence}
\end{equation}
where $\sigma_{\text{trans}} = 0.01\text{ m}$ and $\sigma_{\text{rot}} = 0.1\text{ rad}$ set the characteristic physical scales for translation and rotation jitter, respectively. Given a dataset containing $S$ distinct sequences, the final unified Stability Index reported in our results is computed as the arithmetic mean across all sequences, i.e., $\text{SI} = \frac{1}{S} \sum_{j=1}^{S} \text{SI}_j$. A higher SI indicates a smoother and more stable tracking trajectory.

\subsection{Geometric \& Physical Constraints}
\label{subsec:constraints_hypers}

To complement the hierarchical, dual-stage geometric optimization framework outlined in the main text, this section provides the detailed mathematical formulations, objectives, and parameters for both alignment stages. 

In the first stage, which rectifies systemic global biases and depth-scale ambiguities, the global optimization variables are defined as the scale factor $s \in \mathbb{R}$ and the rigid transformation offsets $\Delta \mathbf{T}_g \in \mathbb{R}^3, \Delta \mathbf{R}_g \in \mathrm{SO}(3)$. To prioritize functional grasping regions, we employ an anatomically weighted contact loss $\mathcal{L}_{\text{c}, t}$ at frame $t$ that emphasizes key sites such as the thumb and index finger:
\begin{equation}
\mathcal{L}_{c, t} = \sum_{f \in \text{Fingers}} w_f \cdot \Phi \Big( \text{dist}(v_{f, t}, \mathcal{O}_t); d_{\text{pen}}, d_{\text{sep}} \Big)
\end{equation}
where $v_{f, t}$ denotes the globally transformed hand keypoints at frame $t$, $\mathcal{O}_t$ is the object point cloud, and $\Phi(\cdot)$ is a penalty function with distance thresholds $d_{\text{pen}}$ and $d_{\text{sep}}$ representing limits for penetration and separation, respectively. The global objective $\mathcal{L}_{\text{g}}$ is minimized across all contact frames $\mathcal{F}_c$ using the L-BFGS-B optimizer:
\begin{equation}
\mathcal{L}_{\text{g}} = \sum_{t \in \mathcal{F}_c} \mathcal{L}_{c, t} + w_{\text{reg}} \Big( \|\Delta \mathbf{T}_g\|_2^2 + \|\log(\Delta \mathbf{R}_g)\|_2^2 + (s-1)^2 \Big)
\end{equation}
where $\log(\cdot)$ maps the rotation matrix to its corresponding Lie algebra $\mathfrak{so}(3)$ (axis-angle representation). This initial stage yields a metric-consistent global trajectory while preserving the relative motion structure of the original tracking.

Building upon the global alignment, we perform fine-grained refinement on a per-frame basis to achieve precise mesh-to-surface alignment. For each contact frame $t \in \mathcal{F}_c$, we optimize a local parameter increment $\delta\Theta = \{\delta\theta_{\text{body}}, \delta\theta_{\text{root}}, \delta\mathbf{t}\}$. The frame-level objective function $\mathcal{L}_t$ integrates the localized contact loss $\mathcal{L}_{\text{c}, t}$ with a temporal smoothness constraint $\mathcal{L}_{\text{s}}$ and a regularization term to prevent excessive deviation from the raw observations:
\begin{equation}
\mathcal{L}_t = \mathcal{L}_{c, t} + w_{\text{sm}}\mathcal{L}_{\text{s}} + w_{\text{reg}}\|\delta\Theta\|_2^2
\end{equation}
For non-contact frames ($t \notin \mathcal{F}_c$), we maintain temporal coherence by propagating the optimization increments from the nearest contact neighbor $t_{n} = \arg\min_{i \in \mathcal{F}_c} |t - i|$. The optimized offset $\Delta \Theta = \Theta^*_{t_n} - \Theta^{\text{init}}_{t_n}$ is propagated to the current frame such that $\Theta^*_t = \Theta^{\text{init}}_t + \Delta \Theta$. This integrated strategy improves temporal consistency and physical feasibility throughout the hand-object interaction sequence. The specific hyperparameter configurations for the core geometric optimization workflow are detailed in Table~\ref{tab:optimization_hyperparameters}.

\begin{table}[htbp]
\centering
\caption{Core Hyperparameter Settings for Geometric Constraints.}
\label{tab:optimization_hyperparameters}
\begin{tabular*}{\linewidth}{@{\extracolsep{\fill}}lcl@{}}
\toprule
\textbf{Hyperparameter} & \textbf{Value} & \textbf{Description} \\
\midrule
Translation Bounds ($\Delta \mathbf{T}_g$) & $\pm 0.05$\,m & Global translation offset bounds. \\
Search Range Factor ($\gamma$) & $0.2$ & Local search bound ratio ($\gamma \cdot \Delta \mathbf{T}_g$). \\
Root Orientation Bounds ($\Delta \mathbf{R}_g$) & $\pm 0.1$\,rad & Global hand rotation offset bounds. \\
Body Pose Bounds ($\delta\theta_{\text{body}}$) & $\pm 0.1$\,rad & Optimization limits for joint angles. \\
\midrule
Finger Weights ($w_f$) & $[40, 20, 20, 10, 10]$ & Weights for fingers in $\mathcal{L}_{c, t}$. \\
Penetration Threshold ($d_{\text{pen}}$) & $0.003$\,m & Proximity/penetration limit in $\Phi(\cdot)$. \\
Separation Threshold ($d_{\text{sep}}$) & $0.001$\,m & Separation margin for active contact. \\
Penetration Penalty ($w_{\text{pen}}$) & $200.0$ & Penalty for mesh intersection. \\
Separation Penalty ($w_{\text{sep}}$) & $10.0$ & Penalty for contact detachment. \\
Regularization Weight ($w_{\text{reg}}$) & $1.0$ & Offset penalty weight \\
Smoothness Weight ($w_{\text{sm}}$) & $1.0$ & Temporal coherence weight for $\mathcal{L}_{\text{s}}$. \\
\midrule
Global Max Iterations ($n_{\text{g\_iter}}$) & $2000$ & Max iterations for global stage. \\
Global Step Size ($\epsilon_{\text{g}}$) & $0.002$ & Gradient step size for global alignment. \\
Global Tolerance ($\text{tol}_{\text{g}}$) & $1\times 10^{-6}$ & Termination tolerance for global phase. \\
\midrule
Per-frame Max Iterations ($n_{\text{l\_iter}}$) & $2000$ & Max iterations for local stage. \\
Refinement Step Size ($\epsilon_{\text{l}}$) & $0.001$ & Gradient step size for local refinement. \\
Per-frame Tolerance ($\text{tol}_{\text{l}}$) & $1\times 10^{-7}$ & Termination tolerance for local phase. \\
\bottomrule
\end{tabular*}
\end{table}

\begin{table}[htbp]
\centering
\caption{Core Hyperparameter Settings for Physical Constraints.}
\label{tab:Physical_hyperparameters}
\begin{tabular*}{\linewidth}{@{\extracolsep{\fill}}lcl@{}}
\toprule
\textbf{Hyperparameter} & \textbf{Value} & \textbf{Description} \\
\midrule
Friction Coefficient ($\mu$) & $0.15$ & Coulomb friction limit for friction cone $\mathcal{FC}_i$. \\
Grasp Energy Weight ($w_{\text{ge}}$) & $100.0$ & Force-closure optimization penalty factor. \\
Distance Loss Weight ($w_{\text{dist}}$) & $1000.0$ & Contact distance objective $\mathcal{J}_{\text{dist}}$ weight. \\
Distance Threshold ($d$) & $0.0$\,m & Target proximity margin in $\mathcal{J}_{\text{dist}}$. \\
Lower Bound Force ($k_{\text{lower}}$) & $0.1$ & Minimum normal force for active contacts. \\
External Disturbance Samples ($j$) & $64$ & Random wrench directions $\mathbf{f}_j \in \mathcal{W}_{\text{ext}}$. \\
QP Solve Interval & $5$ & Step interval for Force-Closure QP solver. \\
\midrule
Optimizer Max Iterations ($N_{\text{k\_iter}}$) & $500$ & Maximum L-BFGS refinement steps. \\
Inner Iterations ($N_{\text{inner}}$) & $50$ & Sub-problem line search iterations. \\
Learning Rate Decay ($\alpha_{\text{decay}}$) & $0.95$ & Optimization step size decay ratio. \\
Wrist Translation Scale ($\eta_{\mathbf{t}}$) & $0.01$ & Base learning rate for wrist position. \\
Wrist Rotation Scale ($\eta_{\mathbf{R}}$) & $0.1$ & Base learning rate for wrist orientation. \\
Joint Pose Scale ($\eta_{\mathbf{q}}$) & $0.1$ & Base learning rate for joint configuration $\mathbf{q}$. \\
Convergence Tolerance ($\epsilon_{\text{conv}}$) & $1\times 10^{-7}$ & Objective function convergence threshold. \\
\bottomrule
\end{tabular*}
\end{table}

For more details regarding the Physical constraints, we present its comprehensive mathematical formulation and execution pipeline below. Specifically, within the Task Wrench Space (TWS), the pipeline samples a set of unit-magnitude random force directions $\mathbf{f}_{j}$ and employs the GJK algorithm to query the object mesh for the nearest contact points and their associated local surface normals $\mathbf{n}_i$:
\begin{equation}
\mathcal{W}_{\text{ext}} = \{ (\mathbf{f}_{j}, \tau_{j}) \mid \| \mathbf{f}_{j} \| = 1, j \in \text{Samples} \}
\end{equation}
For each contact point, we model the friction cone $\mathcal{FC}_i$ based on the GJK-queried normal $\mathbf{n}_i$ and the friction coefficient $\mu$:
\begin{equation}
\mathcal{FC}_i = \{ \mathbf{f}_{c,i} \mid \| \mathbf{f}_{c,i} - (\mathbf{f}_{c,i}^\top \mathbf{n}_i)\mathbf{n}_i \| \leq \mu (\mathbf{f}_{c,i}^\top \mathbf{n}_i) \}
\end{equation}

Finally, we solve for the refined state $\mathbf{x}_t^*$ by minimizing the joint objective $\mathcal{J}$. This formulation optimizes the grasp energy $E_{\text{ge}}$ to ensure that the synthesized wrenches from the friction cones effectively counteract the sampled external disturbances $\mathcal{W}_{\text{ext}}$, while concurrently maintaining a physically consistent contact distance via GJK:
\begin{equation}
\mathcal{J}_{\text{dist}} = w_{\text{dist}} \sum_{i \in \mathcal{C}} (\text{dist}_{\text{GJK}}(\mathcal{M}_{i,t}, \mathcal{O}_t) - d)^2
\end{equation}
\begin{equation}
\min_{\mathbf{x}_t} \mathcal{J} = w_{\text{ge}} E_{\text{ge}}(\mathbf{P}_t, \mathbf{N}_t, \mathcal{W}_{\text{ext}}) + \mathcal{J}_{\text{dist}}
\end{equation}
where $d$ serves as a critical distance threshold that acts as a safety margin to prevent mesh penetration while ensuring close proximity for stable contact. This hierarchical optimization ensures that the final trajectory $\mathcal{X}^* = \{\mathbf{x}^*_t\}_{t=1}^T$ is both kinematically faithful and physically grounded for contact-rich manipulation. The detailed hyperparameter configurations for this Physical workflow are summarized in Table~\ref{tab:Physical_hyperparameters}.

\subsection{Reinforcement Learning Training}
\label{subsec:rl_training}
To supplement the training pipeline and ensure experimental reproducibility, we provide the complete hyperparameter configurations here. Table~\ref{tab:rl_hyperparameters} lists the detailed implementation details for both the Hand Imitator and the Residual Policy, including the PPO optimization parameters and simulation physics configurations. 

Furthermore, Table~\ref{tab:reward_hyperparameters} explicitly breaks down the target quantities, exponential scales ($\alpha$), and composition weights ($w_k$) for each component within the overall reward function. This hyperparameter distribution provides a detailed blueprint of our imitation tracking and contact-driven regularization design.

\begin{table}[htbp]
\centering
\caption{Hyperparameter Configurations for the Hand Imitator and Residual Policy.}
\label{tab:rl_hyperparameters}
\begin{tabular*}{\linewidth}{@{\extracolsep{\fill}}lcc@{}}
\toprule
\textbf{Hyperparameter} & \textbf{Hand Imitator (Base)} & \textbf{Residual Policy} \\
\midrule
Algorithm Name & PPO & PPO \\
Discount Factor ($\gamma$) & $0.99$ & $0.99$ \\
GAE Parameter ($\tau$) & $0.95$ & $0.95$ \\
Clipping ($\epsilon$) & $0.2$ & $0.2$ \\
Learning Rate ($\eta_{\text{RL}}$) & $5\times 10^{-4}$ & $5\times 10^{-4}$ \\
Learning Rate Schedule & Adaptive & Warmup \\
KL Divergence Threshold & $0.008$ & $0.008$ \\
Early Stop Horizon (Epochs) & $500$ & $500$ \\
Horizon Length & $32$ & $32$ \\
Mini-batch Size & $1024$ & $1024$ \\
Mini-epochs  & $5$ & $5$ \\
Gradient Norm Bound & $1.0$ & $1.0$ \\
Value Coefficient ($c_1$) & $4.0$ & $4.0$ \\
Boundary Loss Weight & $1\times 10^{-4}$ & $1\times 10^{-4}$ \\
Actor-Critic MLP Structure & \multicolumn{2}{c}{[256, 512, 128, 64] with ELU Activation} \\
\midrule
Hand Friction  ($\mu_{\text{hand}}$) & $4.0$ & $2.0$ \\
Object Friction ($\mu_{\text{obj}}$) & None & $2.0$ \\
Table Friction ($\mu_{\text{table}}$) & $0.1$ & $0.1$ \\
\bottomrule
\end{tabular*}
\end{table}

\begin{table}[htbp]
\centering
\caption{Hyperparameter Scales and Weights of the Reward Function Components.}
\label{tab:reward_hyperparameters}
\begin{tabular*}{\linewidth}{@{\extracolsep{\fill}}lccc@{}}
\toprule
\textbf{Reward Component} & \textbf{Target Quantity} & \textbf{Scale $\alpha$} & \textbf{Weight $w_k$} \\
\midrule
Wrist Position & $\mathbf{p}_{\text{eef}}$ & $40$ & $0.10$ \\
Wrist Orientation & $\mathbf{R}_{\text{eef}}$ & $1$ & $0.60$ \\
Thumb Tip Position & $\mathbf{p}_{\text{thumb}}$ & $100$ & $0.90$ \\
Index Tip Position & $\mathbf{p}_{\text{index}}$ & $90$ & $0.80$ \\
Middle Tip Position & $\mathbf{p}_{\text{middle}}$ & $80$ & $0.75$ \\
Ring Tip Position & $\mathbf{p}_{\text{ring}}$ & $60$ & $0.60$ \\
Pinky Tip Position & $\mathbf{p}_{\text{pinky}}$ & $60$ & $0.60$ \\
Level-1 Proximal Joints & $\mathbf{p}_{\text{lvl1}}$ & $50$ & $0.50$ \\
Level-2 Distal Joints & $\mathbf{p}_{\text{lvl2}}$ & $40$ & $0.30$ \\
Wrist Linear Velocity & $\mathbf{v}_{\text{eef}}$ & $1$ & $0.10$ \\
Wrist Angular Velocity & $\boldsymbol{\omega}_{\text{eef}}$ & $1$ & $0.05$ \\
Joint Configuration Velocity & $\dot{\mathbf{q}}$ & $1$ & $0.10$ \\
\midrule
Object Position & $\mathbf{p}_{\text{obj}}$ & $80$ & $5.00$ \\
Object Orientation & $\mathbf{R}_{\text{obj}}$ & $3$ & $1.00$ \\
Object Linear Velocity & $\mathbf{v}_{\text{obj}}$ & $1$ & $0.10$ \\
Object Angular Velocity & $\boldsymbol{\omega}_{\text{obj}}$ & $1$ & $0.10$ \\
\midrule
Fingertip Normal Force & $\mathbf{f}_{\text{tip}}$ & $1$ & $1.00$ \\
\bottomrule
\end{tabular*}
\end{table}


\section{Visual Results}
\label{sec:visual_results}
To further demonstrate the robustness and generalization of our proposed framework, this section provides extended visual results and synthesized trajectories that could not be included in the main text due to page constraints.
\subsection{Cross-Embodiment Results}
\label{subsec:cross_embodiment}
\begin{figure}[h] 
    \centering
    \includegraphics[width=\columnwidth]{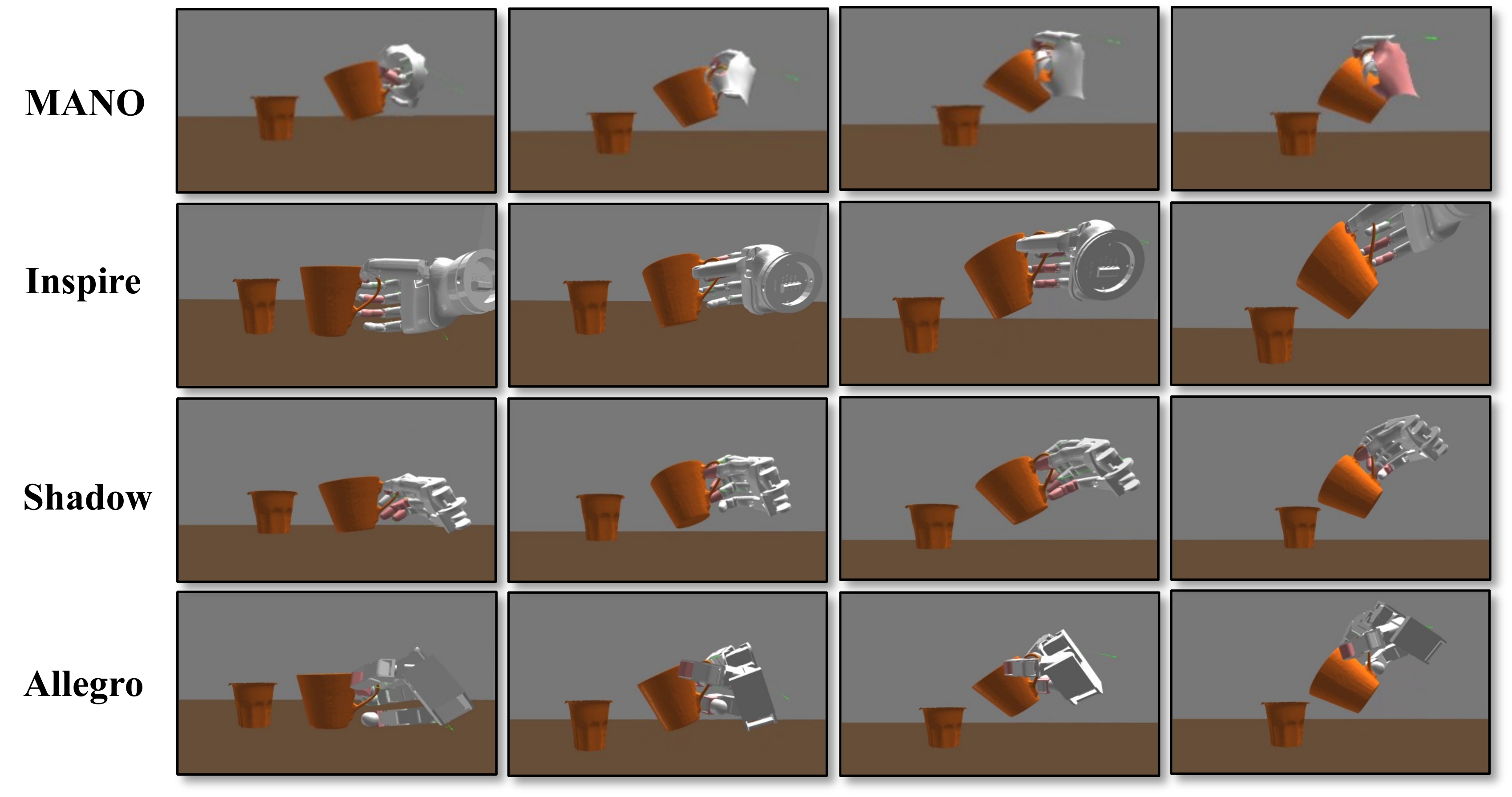} 
    \caption{\textbf{Qualitative results of unimanual manipulation across diverse embodiments.} We showcase single-hand trajectories synthesized from monocular videos across distinct robotic hands.}
    \label{fig:Visual_results_2}
\end{figure}

We train the single-hand policy across $4096$ concurrent environments, reducing to $2048$ for bimanual tasks due to GPU memory limitations. Figure~\ref{fig:Visual_results_2} displays the qualitative trajectories for single-hand manipulation.

\begin{figure}[h] 
    \centering
    \includegraphics[width=\columnwidth]{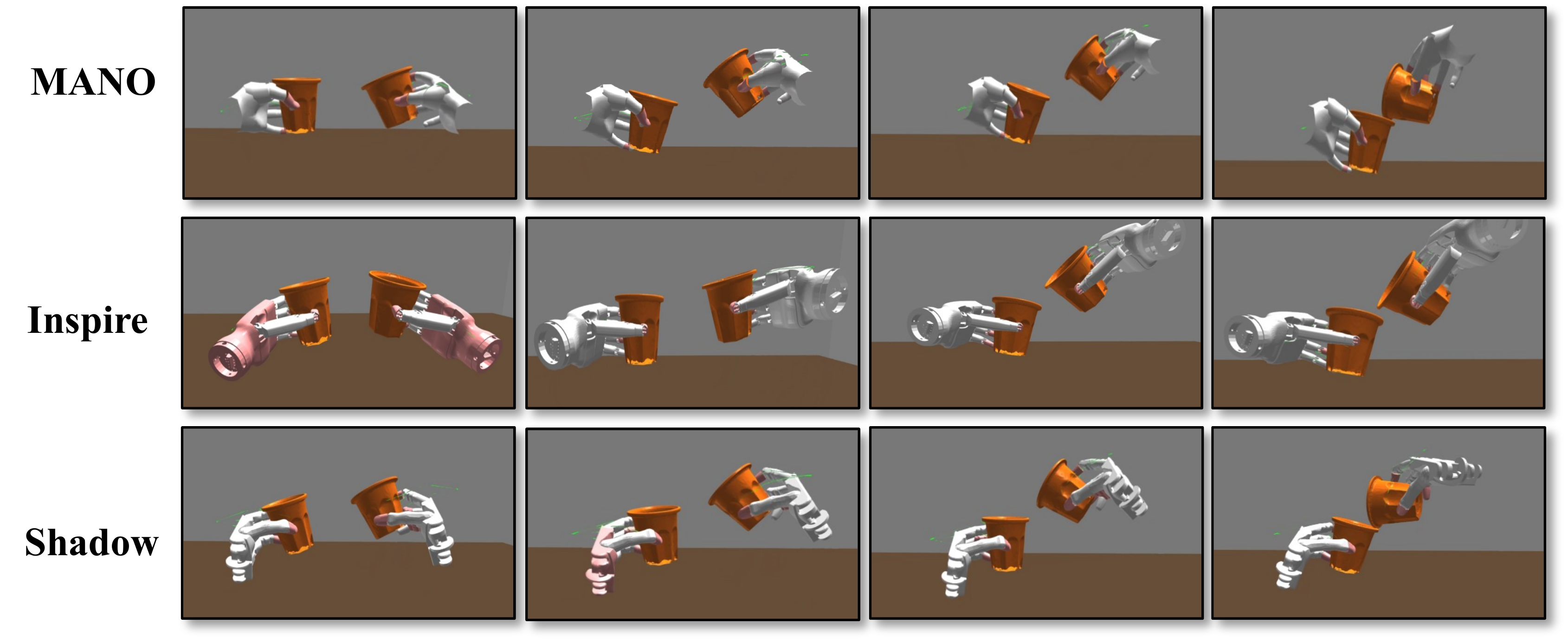} 
    \caption{\textbf{Qualitative results of bimanual manipulation across diverse embodiments.} We extend our pipeline to bi-manual scenarios, visualizing the synthesized trajectories for coordinated dual-hand tasks.}
    \label{fig:Visual_results_3}
\end{figure}

Similarly, Figure~\ref{fig:Visual_results_3} presents the bimanual cooperative tasks. Notably, the bimanual Allegro Hand exhibits degraded performance; due to its bulky link geometry and larger scale, it suffers from joint warping and self-collisions when grasping small-volume objects, highlighting an inherent morphology constraint.

\subsection{Egocentric Results}
\label{subsec:egocentric_results}
\begin{figure}[h] 
    \centering
    \includegraphics[width=\columnwidth]{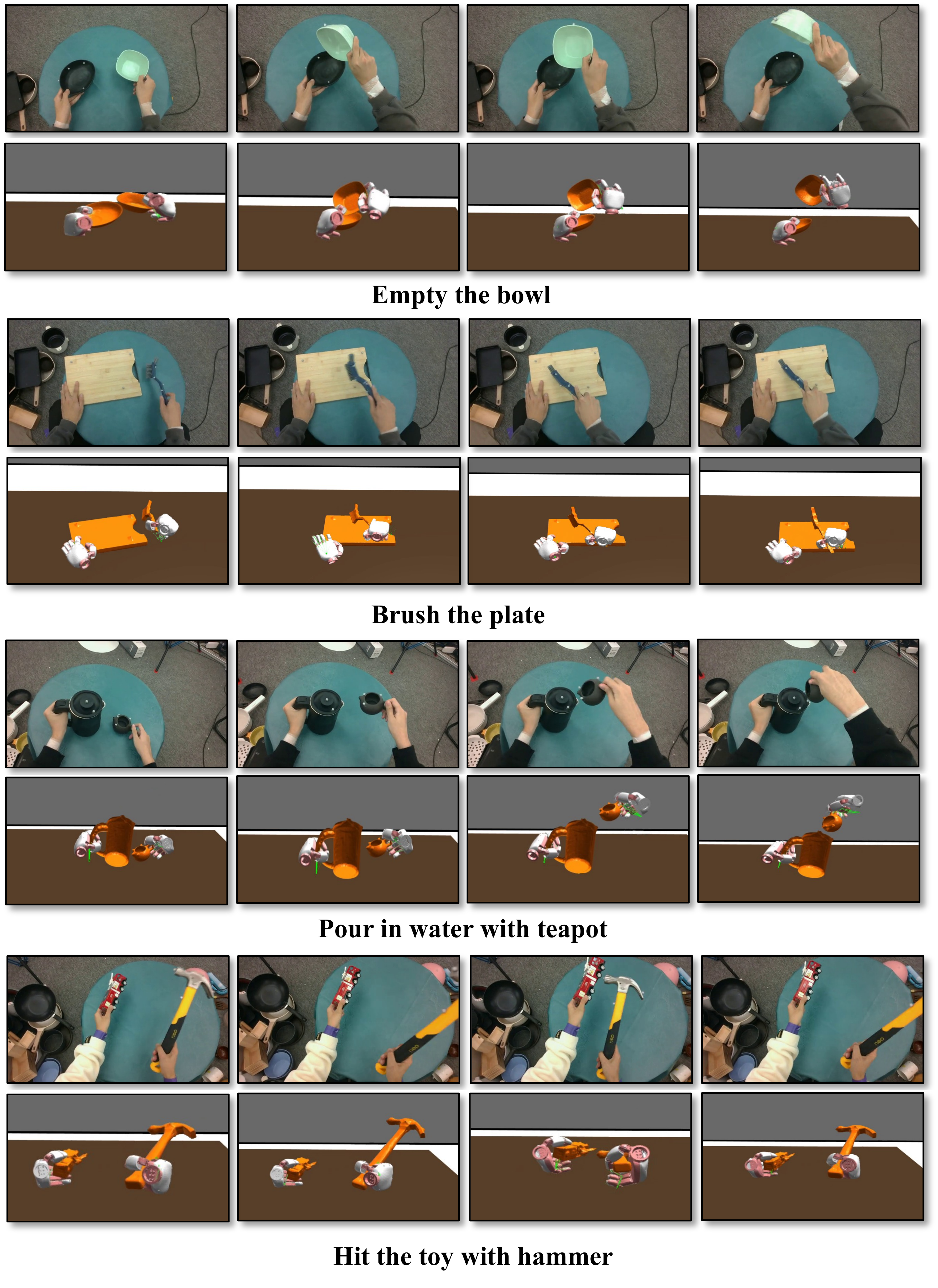} 
    \caption{\textbf{Direct policy learning from egocentric human videos.} We demonstrate that our synthesized trajectories enable the direct learning of robust manipulation policies from raw first-person videos. }
    \label{fig:Visual_results_4}
\end{figure}

To better leverage extensive data sources from existing visual datasets, we directly utilize raw, first-person egocentric videos to synthesize reference motions. As illustrated in Figure~\ref{fig:Visual_results_4}, our framework successfully extracts robust manipulation trajectories from these pervasive video sources, demonstrating its efficacy in turning passive egocentric observations into executable robotic priors.

\subsection{Real-Robot Trajectory Execution}
\label{subsec:real_robot_execution}
To evaluate whether the trajectories recovered from monocular human videos can be deployed on real robotic hardware, we conduct real-robot trajectory execution experiments using a dual-arm UR + XHAND platform. The real system shares the same embodiment as the simulation environment, enabling a direct evaluation of the generated manipulation trajectories.

\begin{figure}[h] 
    \centering
    \includegraphics[width=\columnwidth]{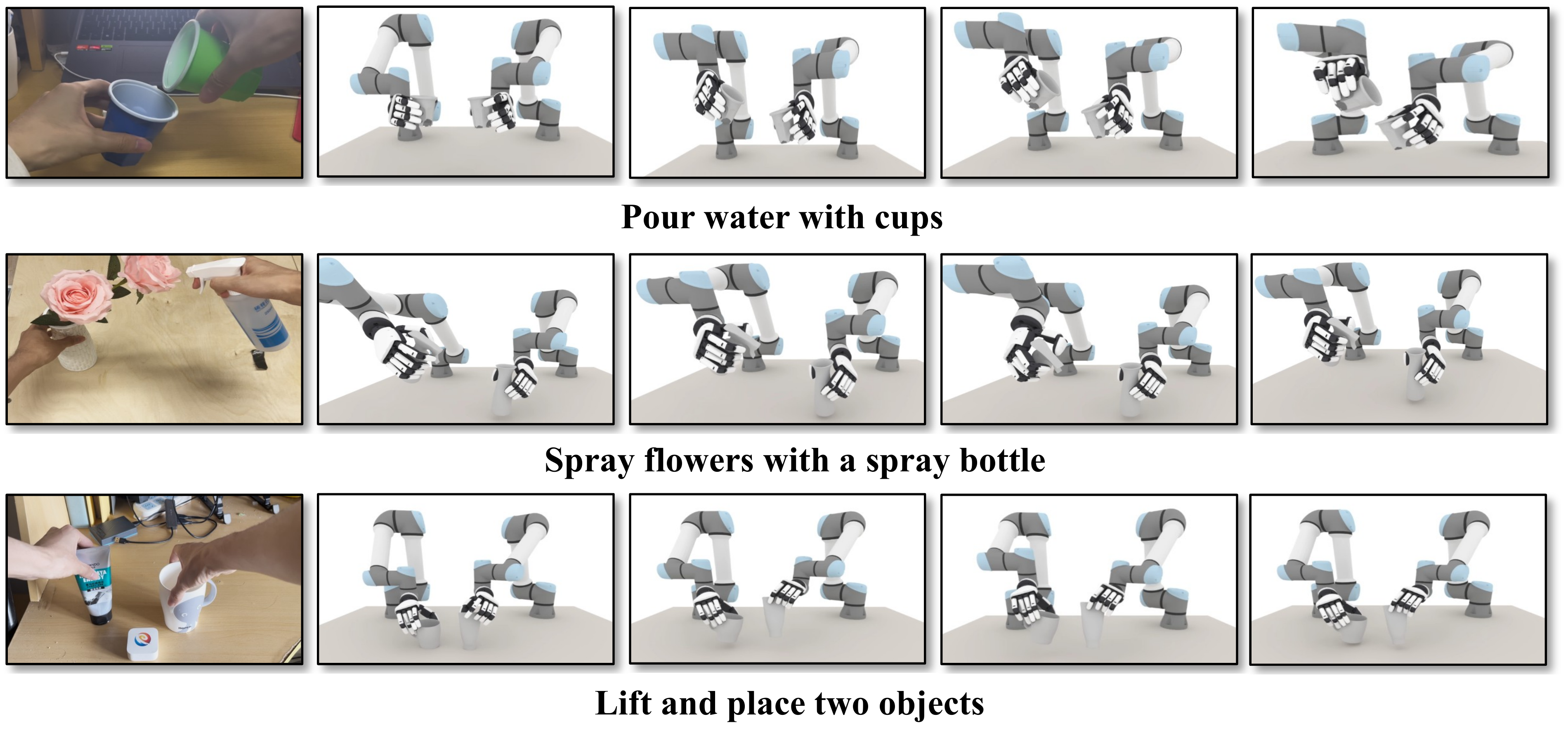} 
    \caption{\textbf{Physics-based execution on the dual-arm UR + XHAND platform.} Video-derived trajectories are transformed into executable arm-hand motions and executed in a physics simulation environment for feasibility verification prior to real-world deployment.}
    \label{fig:Visual_results_5}
\end{figure}
Since the reconstructed demonstrations are represented in a floating-base formulation, they cannot be directly executed on an arm-hand robotic system. To enable deployment on the dual-arm UR + XHAND platform, we first apply our trajectory augmentation pipeline to transform the reconstructed demonstrations into executable arm-hand trajectories. For each generated demonstration, inverse kinematics (IK) is solved throughout the entire trajectory to verify kinematic feasibility. Only trajectories with valid IK solutions are retained for further evaluation. The feasible trajectories are subsequently executed in a physics-based simulation environment to assess task completion and execution stability. Figure~\ref{fig:Visual_results_5} presents representative simulation results on the dual-arm UR + XHAND platform. The trajectories that successfully accomplish the target tasks in simulation are then directly deployed on the real robot without manual intervention.
\begin{figure}[h] 
    \centering
    \includegraphics[width=\columnwidth]{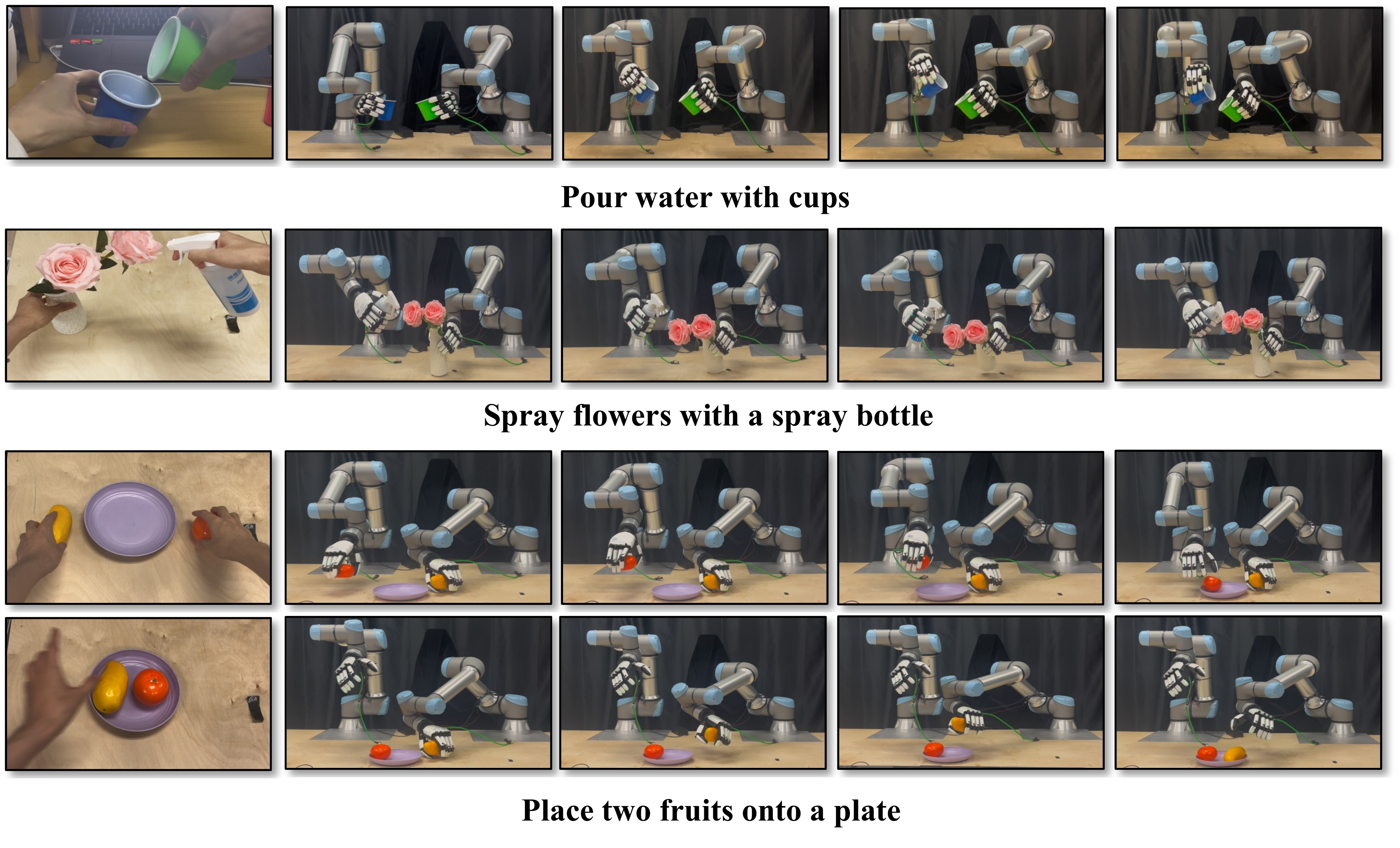} 
    \caption{\textbf{Real-world deployment on the dual-arm UR + XHAND platform.} Trajectories reconstructed from monocular human videos are successfully executed on the real robot, demonstrating effective transfer from video-derived demonstrations to physical manipulation.}
    \label{fig:Visual_results_6}
\end{figure}

Figure~\ref{fig:Visual_results_6} shows representative real-world execution results, where the dual-arm UR + XHAND system successfully performs a diverse set of bimanual manipulation tasks, including pouring water with cups, spraying flowers with a spray bottle, and lifting and placing two objects. The successful deployment of these trajectories on real hardware demonstrates that manipulation behaviors recovered from monocular human videos can be transformed into executable arm-hand motions while preserving the original task intent. Furthermore, the results validate the effectiveness of the proposed geometric and physical constraint optimization framework in producing physically feasible and robot-executable trajectories. Together, these findings demonstrate the ability of our pipeline to bridge the gap between monocular human video demonstrations and real-world dexterous manipulation.

\clearpage
\section{Additional Analysis}
\label{sec:limitations}

In this section, we analyze how Physical constraints facilitate reinforcement learning (RL) training and investigate the root causes of remaining failures. Within the Reference State Initialization (RSI) framework, these constraints are crucial for stabilizing precarious initial grasps. Without them, rough retargeting from monocular video introduces tracking errors that cause severe object penetration or premature detachment. Initializing an RL agent in such physically impossible states leads to irreversible failures and ineffective exploration. To resolve this, we enforce Physical constraints via force-closure optimization, utilizing explicit mesh-to-mesh queries to naturally adapt to complex geometries and ensure physical feasibility across diverse object profiles.

Specifically, we configure a conservative friction coefficient $\mu$ to define a strictly constrained friction cone. This formulation forces the multi-fingered hand to execute more stable grasping motions, enabling the policy to generalize seamlessly to diverse materials and unknown friction conditions. Consequently, the framework rectifies flawed initial configurations into valid, stable contact states, preventing the policy from trapping in pathological local minima, thereby improving exploration efficiency and training convergence. Furthermore, we introduce an additional squeeze phase following the grasp to explicitly compensate for PD controller steady-state errors, ensuring robust force transmission and secure grip maintenance.

However, our framework still exhibits failure cases primarily stemming from the inherent ambiguities of monocular observation. Under an egocentric viewpoint, top-down perspectives heavily compress the vertical dimension, causing the 3D perception module to underestimate object height or thickness. For extremely thin or flat objects like wooden boards, plates, spatulas, and rulers, the reconstructed 3D bounding boxes often possess unphysically low thickness. Given the non-negligible physical thickness of dexterous hands, generating feasible approach trajectories to pinch or scoop these flattened objects directly from a tabletop remains challenging. Our system thus occasionally fails when lifting flat objects off hard surfaces, restricting successful execution to scenarios where the object is pre-elevated or suspended in mid-air.
\end{document}